%% file: naaclhlt2019.tex
\documentclass[11pt,a4paper]{article}
\usepackage[hyperref]{naaclhlt2019}
\usepackage{times}
\usepackage{latexsym}
\usepackage{amsmath}
\usepackage{array,multirow,graphicx}
\usepackage{soul}
\usepackage{subcaption}
\usepackage{url}

\aclfinalcopy 
\def\aclpaperid{1870} 

\title{Giving Attention to the Unexpected:\\ Using Prosody Innovations in Disfluency Detection}

\author{Vicky Zayats and Mari Ostendorf\\
    Electrical \& Computer Engineering Department \\ 
    University of Washington \\
  {\tt [vzayats, ostendor]@uw.edu} \\}

\date{}

\begin{document}
\maketitle
\begin{abstract}
Disfluencies in spontaneous speech are known to be associated with prosodic disruptions. However, most algorithms for disfluency detection use only word transcripts. Integrating prosodic cues has proved difficult because of the many sources of variability affecting the acoustic correlates. This paper introduces a new approach to extracting acoustic-prosodic cues using text-based distributional prediction of acoustic cues to derive vector z-score features (innovations). We explore both early and late fusion techniques for integrating text and prosody, showing gains over a high-accuracy text-only model.
\end{abstract}

\input{intro.tex}
\input{error-anal.tex}
\input{method.tex}

\input{experiments.tex}
\input{analysis.tex}
\input{related.tex}
\input{conclusions.tex}
\input{acknowledgements.tex}

\bibliographystyle{acl_natbib}
\bibliography{disfluency}

\end{document}

%% file: intro.tex
\section{Introduction}

Speech disfluencies are frequent events in spontaneous speech. The rate of disfluencies varies with the speaker and context; one study observed disfluencies once in every 20 words, affecting up to one third of utterances \cite{shriberg94}. Disfluencies are important to account for, both because of the challenge that the disrupted grammatical flow poses for natural language processing of spoken transcripts and because of the information that they provide about the speaker. 

Most work on disfluency detection builds on the framework that annotates a disfluency in terms of a reparandum followed by an interruption point (+), an optional interregnum ($\{ \ \}$), and then the repair, if any. A few simple examples are given below:

{\small
\begin{verbatim}
 [ it's + {uh} it's] almost... 
 [ was it, + {I mean} , did you ] put...
 [I just + I] enjoy working...
 [By + ] it was attached to...
 \end{verbatim}
}
\vskip -0.15in
\noindent
Based on the similarity/differences between the reparandum and the repair, disfluencies are often categorized into three types: repetition (the first example), rephrase (the next example), and restart (the last example).

The interruption point is associated with a disruption in the realization of a prosodic phrase, which could involve cutting words off or elongation associated with hesitation, followed by a prosodic reset at the start of the repair. There may also be emphasis in the repair to highlight the correction.

Researchers have been working on automatic disfluency detection for many years \cite{lickley1994detecting, shriberg1997prosody,charniak01,johnson04,Lease+06,qian+13,zayats2016}, motivated in part by early work on parsing speech that assumed reliable detection of the interruption point \cite{nakatani94,shriberg97,liu06}. The first efforts to integrate prosody with word cues for disfluency detection \cite{baron2002automatic,snover2004lexically} found gains from using prosody, but word cues played the primary role. In subsequent work \cite{qian+13,honnibaljoint,wang2017transition},
more effective models of word transcripts have been the main source of performance gains. The success of recent neural network systems raises the question of what the role is for prosody in future work.
In the next section, we hypothesize where prosody might help and look at the relative frequency of these cases and the performance of a high accuracy disfluency detection algorithm in these contexts.

With the premise that there is a potential for prosody to benefit disfluency detection, we then propose a new approach to extracting prosodic features. A major challenge for all efforts to incorporate prosodic cues in spoken language understanding is the substantial variability in the acoustic correlates of prosody. For example, duration cues are expected to be useful -- disfluencies are often associated with duration lengthening related to hesitation. However, duration varies with phonetic context, word function, prosodic phrase structure, speaking rate, etc. To account for some of this variability, various feature normalization techniques are used, but typically these account for only limited contexts, e.g. phonetic context for duration or speaker pitch range for fundamental frequency.  In our work, we introduce a mechanism for normalization using the full sentence context. We train a sequential neural prediction model to estimate distributions of acoustic features for each word, given the word sequence of a sentence. Then, the actual observed acoustic feature is used to find the prediction error, normalized by the estimated variance. We refer to the resulting features as innovations, which can be thought of as a non-linear version of the innovations in a Kalman filter. The innovations will be large when the acoustic cues do not reflect the expected prosodic structure, such as during hesitations, disfluencies, and contrastive or emphatic stress. The idea is to provide prosodic cues that are less redundant with the textual cues.
We assess the new prosodic features in experiments on disfluency detection using the Switchboard corpus, exploring
both early and late fusion techniques to integrate innovations with text features. Our analysis shows that prosody does help with detecting some of the more difficult types of disfluencies.

This paper has three main contributions. First, our analysis of a high performance disfluency detection algorithm confirms hypotheses about contexts where text-only models have high error rates. Second, we introduce a novel representation of prosodic cues, i.e. the innovation vector resulting from predicting prosodic cues given the whole sentence context. Analyses of the innovation distributions show expected patterns of prosodic cues at interruption points. Finally, we demonstrate improved disfluency detection performance on Switchboard by integrating prosody and text-based features in a neural network architecture, while comparing early and late fusion approaches. 


%% file: error-anal.tex
\section{How Might Prosody Help?}
\label{sec: error-anal}

\begin{table} [t]
\begin{center}
\begin{tabular}{|l|c|c|c|c|c|c|} \hline
            & \multicolumn{4}{c|}{\bf  Reparandum Length} & \bf \% in\\ \cline{1-5}
\bf Type    &  \bf 1-2  &\bf 3-5 & \bf 6-8 & \bf 8$+$  & \bf type \\ \hline
repetition  & 1894 & 419 & 12 & 1 & 46\%\\ \hline
rephrase  & 794 & 585 & 66 & \textendash & 28\%\\ \hline
restart  & 196 & 14 & \textendash & \textendash & 4\%\\ \hline
nested* & 149 & 262 & 158 & 118 & 13\%\\ \hline
\end{tabular}
\end{center}
\caption{\label{tab:disfl_counts} Total word counts associated with reparanda of different lengths and types of disfluencies. *Counts for nested disfluencies exclude repetition tokens.}
\end{table}

\begin{table} [t]
\begin{center}
\begin{tabular}{|l|c|c|c|c|c|} \hline
            & \multicolumn{4}{c|}{\bf  Reparandum Length} & \\ \cline{1-5}
\bf Type    &  \bf 1-2  &\bf 3-5 & \bf6-8 & \bf 8$+$  & \bf overall \\ \hline
repetition  & 0.99 & 0.99  & 1 & 1 & 0.99 \\ \hline
rephrase      & 0.75 & 0.66  & 0.44 & \textendash & 0.70 \\ \hline
restart     & 0.41 & 0  & \textendash & \textendash & 0.39 \\ \hline
$\text{nested}^{*}$   & 0.79 & 0.66  & 0.62 & 0.21 & 0.62 \\ \hline
\end{tabular}
\end{center}
\caption{\label{tab:error_analysis} Percent of reparandum tokens that were correctly predicted as disfluent.  *Statistics for nested disfluencies exclude repetition tokens.}
\end{table}

\begin{table} [t]
\begin{center}
\begin{tabular}{|l|c|c|} \hline
            & \multicolumn{2}{c|}{\bf  Reparandum Length} \\ \hline
\bf Type    &\bf 1-2  &\bf 3-5 \\ \hline
content-content       & 0.61 (30\%) & 0.58 (52\%)  \\ \hline
content-function       & 0.77 (20\%) & 0.66 (17\%) \\ \hline
function-function       & 0.83 (50\%) & 0.80 (32\%) \\ \hline
\end{tabular}
\end{center}
\caption{\label{tab:cw-fw} Relative frequency of rephrases correctly predicted as disfluent for disfluencies that contain a content word in both the reparandum and repair (content-content), either the reparandum or repair (content-function) or in neither. Percentages in parentheses show the fraction of tokens belong to each category.}
\end{table}

Disfluency detection algorithms based on text alone rely on the fact that disfluencies often involve parallel syntactic structure in the reparandum and the repair, as illustrated in the previous examples. In these cases, pattern match provides a strong cue to the disfluency. 
In addition, ungrammatical function word sequences are frequently associated with disfluencies, and these are relatively easy for a text-based model to learn.
In some cases, an interregnum word (or words) provides a word cue to the interruption point. 
In the Switchboard corpus, only 15\% of interruption points are followed by an interregnum, but it can provide a good cue when present. 
Prosody mainly serves to help identify the interruption point.
Thus, for these types of disfluencies, it makes sense that prosodic cues would not really be needed.

Because disfluencies with a parallel syntactic structure do represent a substantial fraction of disfluencies in spontaneous speech, text-based algorithms have been relatively effective. The best models achieve F-scores of 86-91\%\footnote{It is difficult to directly compare published results, because there are different approaches to tokenization that have a non-trivial impact on performance but are not well documented in the literature. Those differences include handling of fragment words, turn boundaries, and tokenization. For example, some studies use fragment features explicitly, while others omit them because speech recognition systems often miss them. Turn boundaries that do not end with a slash unit pose an ambiguity during speaker overlap: cross-turn 'sentences' can either be combined into a longer sentence or separated based on the turn boundary, which impacts what can be detected. Lastly, there are differences in whether contractions and possessives are split into two tokens, and whether conversational terms such as ``you know'' are combined into a single token.}
\cite{lou2017disfluency,zayats2018robust,wang2017transition,wang2018semi}.
We hypothesize that many errors are associated with contexts where we expect that prosodic cues are useful, 
specifically the five cases below, with examples from the development set.

\noindent
{\bf Restarts:} Some disfluencies have no repair; the speaker simply restarts the sentence with no obvious parallel phrase. 

{\small
\begin{verbatim}
 [ it would be + ] I think it's clear...
 well [the +] uh i think what changed...
\end{verbatim} 
}

\noindent
{\bf Long disfluencies:} These include distant pattern match or substantial rephrasing. 

{\small
\begin{verbatim}
 [there is + for people who don't want 
to do the military service it would be 
neat if there were]
 [what they're basically trying to do + 
i don't know up here in massachusetts 
anyhow what they're basically trying to 
do]
\end{verbatim} 
}

\noindent
{\bf Complex (nested) disfluencies:} Disfluencies can occur within other disfluencies. 

{\small
\begin{verbatim}
 [really + [[i + i] + we were really]...
 [[to + to try to] + for two people who 
don't really have a budget to] ]...
\end{verbatim} 
}

\noindent
{\bf Non-trivial rephrasing:} Rephrasing does not always involve a simple ``rough copy" of a repair.

{\small
\begin{verbatim}
 [can + still has the option of]...
 to keep them [in + uh quiet ]...
\end{verbatim} 
}

\noindent
{\bf Fluent repetitions:} Contexts with fluent repetitions often include expressing a strong stance. 

{\small
\begin{verbatim}
 a long long time ago...
 she has very very black and white... 
 \end{verbatim}
}
\vskip -.12in
In order to confirm that there is potential for prosody to help in these contexts, we first
categorize the disfluencies.
To avoid hand-labeling of categories, we distinguished disfluencies based on surface forms (repetition, rephrase, restart) and length of the disfluency reparandum.  Word counts for the different categories are given in Table~\ref{tab:disfl_counts}.

Conditioning on the different contexts, we analyze errors in the development set
made by the high accuracy text-based disfluency detection system that is the baseline for this study \cite{zayats2018robust}. For this model, trained on Switchboard, the performance is 87.4 F-score (P=93.3, R=82.2) on the development set and 87.5 (P=93.1, R=82.5) on the test set.
For each class, we measured the disfluency detection recall (relative frequency of reparandum tokens that were predicted correctly), as well as the percentage of tokens associated with each class. The results in Table \ref{tab:error_analysis} confirm that error rates are higher for restarts, longer rephrasings, and complex disfluencies. 

Rephrase disfluencies include both short lexical access errors, as well as non-trivial rewordings, which tend to be longer and involve content words. Table \ref{tab:cw-fw} breaks down performance for different lengths and word class to explore this difference. We found that rephrase disfluencies that contain content words are harder for the model to detect, compared to rephrases with function words only, and error increases for longer disfluencies. 

Finally, the relative frequency of false positives in fluent repetitions is 0.35. Since
fluent repetitions account for only 4\% of all repetitions, the impact on overall performance is small.

The ultimate goal of a disfluency detection system is to perform well in domains other than Switchboard. Other datasets are likely to have different distributions of disfluencies, often with a higher frequency of those that are hard to detect, such as restarts and repairs \cite{zayats2014multidomain}. In addition, due to the differences in vocabulary, disfluencies with content words are more likely to get misdetected if there is a domain mismatch.  Thus, we hypothesize that prosody features can have a greater impact in a domain transfer scenario. 

%% file: method.tex
\section{Method}
\begin{figure*}[t]
    \centering
    \begin{subfigure}[b]{0.37\textwidth}
        \centering
        \includegraphics[width=\textwidth]{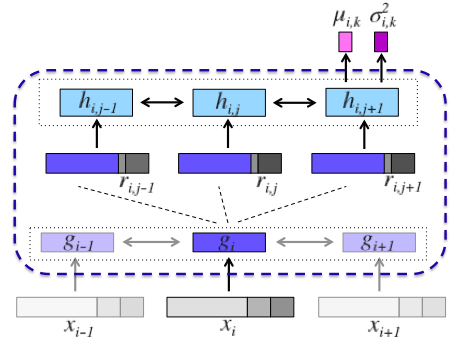}
        \caption{Prosody prediction model}
    \label{fig: prosody_prediction}
    \end{subfigure}%
    \quad
    \begin{subfigure}[b]{0.6\textwidth}
        \centering
        \includegraphics[width=\textwidth]{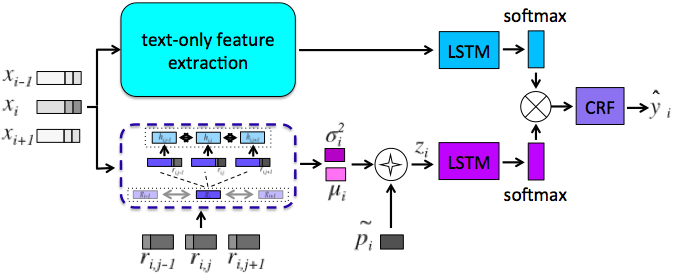}
        \caption{Late fusion model}
    \label{fig: late_fusion}
    \end{subfigure}
    \caption{Prosody prediction (\textbf{left}) and late fusion (\textbf{right}) models. $x_i$ is a contcatenation of token, POS and identity features embeddings at time $i$; $r_i,j$ is a concatenation of stress and phone embeddings for phone $j$ in token $i$; $\widetilde{p_i}$ is a vector of prosodic cues; $g_i$ and $h_i$ are hidden states of token level and phone level LSTMs, correspondingly. }
\end{figure*}

Integrating prosodic cues has proved difficult because of the many sources of variability affecting the acoustic correlates, while systems that only use text achieve high performance. In this work, we propose a new approach that operates on differences in information found in text and prosody. In order to calculate such differences, we introduce innovation features, similar to the concept of innovations in Kalman filters. The key idea is to predict prosodic features based on text information, and then use the difference between the predicted and observed prosodic signal (innovations) as a new feature that is additionally used to predict disfluencies. 

Let a prosody cue, $p_i$ at time $i$ be an observation associated with a sentence transcript containing $n$ word tokens, $x_0\dots x_n$. 
This observation can be modeled as a function of the sentence context $H(x_0\dots x_n)$ perturbed with Gaussian noise $v_i \sim \mathcal{N}(0,\,\sigma^2_{i})$:
\begin{equation}
    p_i = H(x_0\dots x_n) + v_i
\end{equation}
$v_i$ can be viewed as a difference in information found between text and prosody. This difference can be measured using a z-score, 
which is a measure of how many standard deviations below or above the population mean an observation is. 
This framework can be viewed as a non-linear extension of a Kalman filter, where both $H$ and $\sigma^2_{i}$ are parametrized using a neural network. Since disfluencies are irregularities in spoken language, they can be considered anomalies to fluent speech flow. 
A prosody flow that is unusual for a given word sequence, such as one that happens at interruption points,
will likely have higher deviation from the predicted distribution. This anomaly in speech flow provides a strong signal when extracted using innovations, which is complementary to the text cues. 
In the next sections we give more details about the neural network architecture for text encoding, prosodic cues and innovation features, as well as an overview of the whole system. 

\subsection{Text Encoding for Prosody Prediction}
We use both context around a word as well as subword information in text encoding for prosody prediction. Our text encoding consists of two bidirectional LSTMs: one on the token level and another on the phone level. First, we use pretrained word embeddings \cite{levy2014dependency}, part-of-speech tags embeddings, and identity features (whether the word is a filled pause, discourse marker, or incomplete) as inputs to a word-level bidirectional LSTM. Then, for each phone in a word we concatenate the phone embedding, its stress embedding, and the hidden state of the word-level LSTM for the corresponding token. The resulting phone feature vector is used as input to the second bidirectional LSTM. The last hidden state $h_i$ of this second LSTM for token $i$ summarizes the phone, stress and context information of that token, which we use to predict word-level prosodic cues. We use 3 categories of stress features in our experiments: primary, secondary and a non-stress phone.

\subsection{Prosodic Cues}
\label{sec:prosody_feats}

Our prosodic cues include:

\noindent \textbf{Pause.}
Given a pause before a word, $r_i$, our pause cues are scaled as follows:
\begin{equation}
\widetilde{r_i} = \min(1, \ln{(1 + r_i)})    
\end{equation}
Pause information is extracted on a word-level using Mississippi State (MsState) time alignments (more details on data preprocessing in Section \ref{sec: data}.) 
We use scaled real-valued pause information. Scaling pause lengths this way, including the threshold for pauses longer than 1 sec (which are rare), makes the pause distribution less skewed. 

\noindent \textbf{Word Duration.}
Similar to pause information, we extract word duration information using MsState time alignments. 
We do not need to do the standard word-based duration normalization, since the idea behind the innovation model is to normalize prosodic features using a richer context representation.

\noindent \textbf{Fundamental frequency (F0) and Energy (E).} 
Similar to \citet{tran2017joint}, we use three F0 features and three energy features. The three F0 features include normalized cross correlation function (NCCF), log-pitch weighted by probability of voicing (POV), and the estimated delta of log pitch. The three energy features include the log of total energy, the log of total energy from lower 20 mel-frequency bands and the log of total energy from higher 20 mel-frequency bands. The contour features are extracted from 25-ms frames with 10-ms hops using Kaldi \cite{kaldi}. 
Our model is trained to predict the mean of these features across the frames in a word.

\noindent \textbf{MFCCs.} In addition to features used in \citet{tran2017joint}, we also use 13 mel-frequency cepstral coefficients, averaged at the word level, similar to F0 and energy features as described above.

\newcommand{\STAB}[1]{\begin{tabular}{@{}c@{}}#1\end{tabular}}

\begin{table*} [t]
\begin{center}
\begin{tabular}{|l|l|c|c|c|c|c|} \hline
& \multicolumn{1}{l|}{\bf Model} & \multicolumn{2}{c|}{\bf dev} & \multicolumn{2}{c|}{\bf test} & \multicolumn{1}{c|}{\textbf{$\alpha$}} \\
&  & mean & best & mean & best & \\ \hline
\parbox[t]{2mm}{\multirow{3}{*}{\STAB{\rotatebox[origin=c]{90}{single}}}} 
& text                 & 86.54 & 86.80 & 86.47 & 86.96 & \textendash \\
& raw                  & 35.00 & 37.33 & 35.78 & 37.70 & \textendash \\
& innovations          & 80.86 & 81.51 & 80.28 & 82.15 & \textendash \\ \hline
\parbox[t]{2mm}{\multirow{3}{*}{\STAB{\rotatebox[origin=c]{90}{early}}}} 
& text + raw           & 86.46 & 86.65 & 86.24 & 86.53 & \textendash \\ 
& text + innovations       & 86.53 & 86.77 & 86.54 & 87.00 & \textendash \\ 
& text + raw + innovations & 86.35 & 86.69 & 86.55 & 86.44 & \textendash \\ \hline
\parbox[t]{2mm}{\multirow{3}{*}{\STAB{\rotatebox[origin=c]{90}{late}}}} 
& text + raw           & 86.71 & 87.05 & 86.35 & 86.71 & 0.2 \\ 
& text + innovations       & \bf 86.98 & \bf 87.48 & \bf 86.68 & \bf 87.02 & 0.5 \\ 
& text + raw + innovations & 86.95 & 87.30 & 86.60 & 86.87 & 0.5 \\ \hline
\end{tabular}
\end{center}
\caption{\label{tab:results}F1 scores on disfluency detection when using a single set of features (text-only, raw prosody features or innovation features), with early fusion and late fusion. ``Raw'' indicates the usage of original prosodic features (Section \ref{sec:prosody_feats}), while ``innovations'' indicate the usage of innovation features (Section \ref{sec:anomaly_feats}).}
\end{table*}

\subsection{Prosody Innovation Cues}
\label{sec:anomaly_feats}

Given a word-level text encoding $h_i$, for each token in a sentence we predict each of the $k$ prosodic cues $\widetilde{p_i}^k$ listed above. We assume that the predicted prosody cues conditioned on text have a Gaussian distribution:
\begin{align}
\begin{split}
\widetilde{p_i}^k|&h_i \sim \mathcal{N}(\mu_{i,k},\,\sigma^2_{i,k}) \\
\mu_{i,k} &= f(W_1^k h_i + b_1^k) \\
\sigma^2_{i,k} &= softplus(W_2^k h_i + b_2^k)
\end{split}
\end{align}
$W_1^k$, $b_1^k$, $W_2^k$, $b_2^k$ are learnable parameters; the activation function 
$$softplus(x) = \log(1 + \exp(x))$$ 
ensures that the variance is always positive; $f$ is an activation function, which is $softplus$ for pauses and durations, and $tanh$ for the rest of the prosodic cues. The objective function is a sum of the negative log-likelihood of prosodic cues $\widetilde{p_i}^k$ given text encoding.
Then, given the predicted $\mu_{i,k}$, $\sigma^2_{i,k}$ and true values of prosodic cues $\widetilde{p_i}^k$, we calculate z-scores 
for each of the cues, which should have high absolute value for tokens with unusual prosodic behaviour:
\begin{equation}
 z_i^k = \frac{\widetilde{p_i}^k - \mu_{i,k}}{\sigma_{i,k}}
\end{equation}
The prosody prediction module is illustrated in Figure \ref{fig: prosody_prediction}. 

These z-scores, or \emph{innovations}, are used as additional features in our disfluency detection model. 
We train the prosody prediction model only on sentences that do not contain any disfluencies. Any unusual behaviours in disfluency regions, therefore, should have large innovation values predicted by our model.

\subsection{Disfluency Detection System}
Following \cite{zayats2018robust}, we use a bidirectional LSTM-CRF model as our disfluency detection framework. This framework uses a BIO tagging approach, where we predict whether each token is a part of a reparandum, repair or both. 
Following previous studies, the overall performance is measured in F-score of correctly predicted disfluencies in the reparandum. Previous work used textual features only. Here, we evaluate the importance of innovation cues with two types of multimodal fusion - early and late fusion. In early fusion, we concatenate innovations and/or prosody features with the rest of the textual features used in the framework at the input to LSTM layer. In late fusion, we create two separate models - one with only textual features and another with innovations and/or prosody features. Then we do a linear interpolation of the states of two models just before feeding the result to the CRF layer:
\begin{equation}
\label{eq: alpha}
    u_i^{shared} = \alpha u_i^{prosody} + (1 - \alpha) u_i^{text}
\end{equation}
We tune the interpolation weight $\alpha$ and report the best in our experiments section. We train our model jointly, optimizing both prosodic cues prediction and disfluency detection. The schematic view of the late fusion system is presented in Figure \ref{fig: late_fusion}.

%% file: experiments.tex
\section{Experiments}
In our experiments we evaluate the usefulness of innovation features, and compare it to baselines with text-only or raw prosodic cues. 
For each model configuration, we run 10 experiments with different random seeds. This alleviates the potential of making wrong conclusions due to ``lucky/unlucky'' random seeds. We report both the mean and best scores among the 10 runs. 

\begin{table*} [t]
\begin{center}
\begin{tabular}{|l|} \hline
 but it 's just you know {\color{red}leak leak} leak everywhere \\ 
 people	should	know	{\color{red} that }	that	's	an	option \\
 and	i	think	you	{\color{red}do}	accomplish	more	after	that \\
 i	mean [ {\color{red} it	was }	+ it ] \\
 interesting	thing [ {\color{red} about	gas	is	when }	+ i	mean	about	battery	powered	cars	is ] \\ 
\hline
\end{tabular}
\end{center}
\caption{\label{tab: examples_better} Examples of sentences where prosody innovations help. Words in red are correctly labeled when using prosody but not with text only. The first three show fluent phrases; the last two have disfluencies that are missed without prosody.}
\end{table*}

\begin{table*} [t]
\begin{center}
\begin{tabular}{|l|} \hline
i	like	to	run	[about +	oh	about ]	[{\color{red}two} + two	and	a	half ]	miles \\
{\color{red} the	old-timers}	even	the	people	who	are	technologists	do	n't	know	how	to	operate \\
 i	do	n't	know	whether	that	's	because	{\color{red} they}	you know	sort	of	give	up	hope \\
 it	must	be	really	{\color{red}challenging	to}	um	try	to	juggle	a	job \\
\hline
\end{tabular}
\end{center}
\caption{\label{tab: examples_worse} Examples of the sentences where prosody innovations hurt. Words in red are incorrectly labeled when using prosody but not with text only. The first shows a disfluency missed when using prosody; the other three are fluent regions with false detections.}
\end{table*}

\subsection{Data Preprocessing}
\label{sec: data}
Switchboard \cite{Switchboard} is a collection of telephone conversations between strangers, containing 1126 files  hand-annotated with disfluencies. Because human transcribers are imperfect, the original transcripts contained errors. MsState researchers ran a clean-up project which hand-corrected the transcripts and word alignments \cite{deshmukh1998resegmentation}. In this work, we use the MsState version of the word alignments, which allows us to extract more reliable prosodic features. Since the corrected version of Switchboard does not contain updated disfluency annotations, we corrected the annotations using a semi-automated approach: we used a text-based disfluency detection algorithm to re-annotate tokens that were corrected by MsState, while keeping the rest of the original disfluency annotations. The result is referred to as a silver annotation. Most of the corrected tokens are repetitions and restarts. 
To assess the quality of the automatic mapping of disfluencies, we hand-annotated a subset (6.6k tokens, 453 sentences) of the test data and evaluated the performance of the silver annotation against the gold annotation, which has an F1 score of 90.1 (Prec 90.1, Rec 90.1).
Comparing the performance estimates from gold and silver annotations on this subset, we find that the silver annotations give somewhat lower F1 scores (2-3\% absolute), both due to lower precision and recall scores.

\subsection{Results}

Our experiments evaluate the use of innovations with two popular multimodal fusion approaches: early fusion and late fusion.
Our baselines include models with text-only, prosody cues only (raw), and innovation features only as inputs. Since innovations require both text and raw prosodic cues, this baseline is multimodal. In addition, for the late fusion experiments, we show the optimal value of $\alpha$, the interpolation weight from Equation \ref{eq: alpha}. All experiment results are presented in Table \ref{tab:results}. 

We found that innovations are helpful in both early and late fusion frameworks, while late fusion performs better on average. The interpolation weight $\alpha$ for the late fusion experiments is high when innovations are used, which further indicates that innovation features are useful in overall prediction. Interestingly, innovation features alone perform surprisingly well. We also take a closer look at the importance of joint training of the disfluency detection system with prosody prediction. To do this, we pretrain the prosody prediction part of the model first. Then, we train the full model with innovation inputs while freezing the part of the network responsible for predicting prosodic cues.
The mean F-score of this disjointly trained model is 49.27\% on the dev set, compared to 80.86\% for the jointly trained model. This result suggests that training the system end-to-end in a multitask setup is very important. 

%% file: analysis.tex
\section{Analysis}

\subsection{Error analysis}

In order to better understand the impact of the prosody innovations, we perform an error analysis where we compare the predictions of two models: a late fusion model that uses both text and innovation features, and a baseline model that uses text only. All of the analysis is done on the dev set with the model that has the median performance out of 10 that were trained. 

First, we extract all the sentences where the number of disfluency detection errors using the innovation model is lower than when using the text-only model (168 sentences). Examples of such sentences are presented in Table \ref{tab: examples_better}. By looking at the sentences where the model with innovations performs better, we see fluent repetitions and other ambiguous cases where audio is useful for correctly identifying disfluencies.  

On the other hand, in Table \ref{tab: examples_worse}, we have examples of sentences that have a higher number of errors when prosody is used (143 sentences). In the first example, the labeling of ``two'' as fluent by the model with prosody is arguably correct, with the repetition indicating a range rather than a correction. The next  involves a parenthetical phrase, the start of which may be confused with an interruption point. In the last two cases, there is a prosodic disruption and an interegnum, but no correction. 


In order to understand whether incorporating prosody through our model supports the hypotheses in Section \ref{sec: error-anal}, we compare the performance of two models for different categories of disfluencies. We found that using prosody innovations improves detection of:
non-repetition disfluencies (from 68.2\% to 73.7\%), particularly for disfluencies with content words (65.2\% to 71.0\%);
 long repairs (64.0\% to 72.7\% and 40.0\% to 64.6\% for disfluencies with length of repair greater than 3 and 5 correspondingly); and
restarts (from 36.0\% to 37.4\%).
Prosodic innovations also help decrease the rate of false positives for fluent repetitions: the false positives rate decreased from 46.5\% to 38.4\%. 
However, the prosody model increases the false positives in other contexts, such as in the examples in Table~\ref{tab: examples_worse}.



\subsection{Innovation Predictors}


\begin{figure}[t]
    \centering
  \includegraphics[width=\columnwidth]{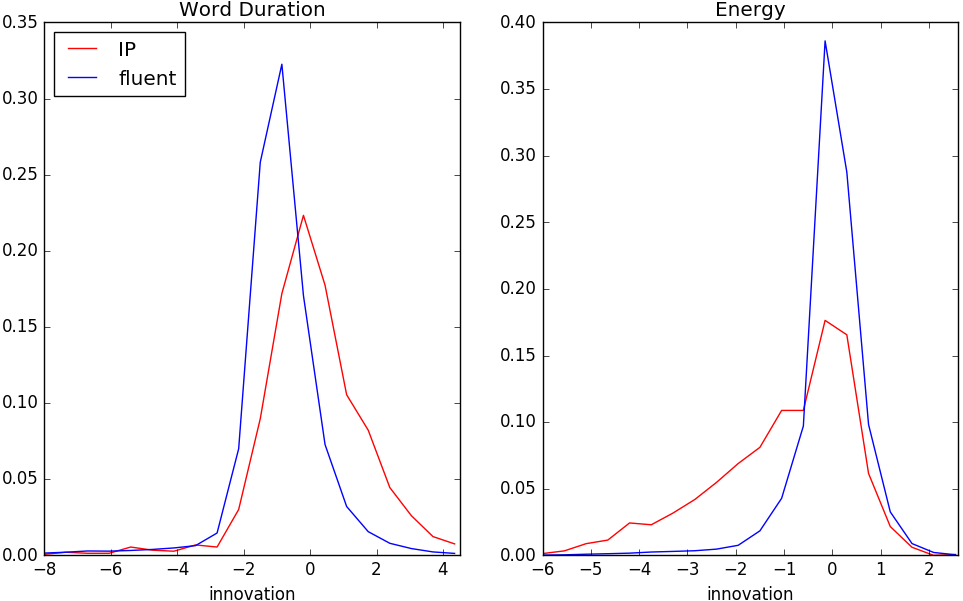}
\caption{Histogram of innovations for word duration and energy features for words preceding an interruption point vs.\ fluent words.}
\label{fig: histogram}
\end{figure}

In order to understand what the model actually learns with respect to innovations, we look at innovation distributions for words preceding interruption points compared to fluent words. The histograms are presented in Figure \ref{fig: histogram}. As expected, we see that words preceding interruption points have atypically longer duration and lower energy. The intonation features did not show substantial distribution differences, probably due to the overly simplistic word-level averaging strategy.


%% file: related.tex
\section{Related Work}

Most work on disfluency detection falls into three main categories: sequence tagging, noisy-channel and parsing-based approaches. Sequence tagging approaches rely on BIO tagging with recurrent neural networks \cite{disfluency_rnn,zayats2016,wang2016neural,zayats2018robust,lou2018disfluency}. Noisy channel models operate on a relationship between the reparandum and repair for identifying disfluencies \cite{charniak01,zwarts+10}. \citet{lou2017disfluency} used a neural language model to rerank sentences using the noisy channel model. 
Another line of work combined parsing and disfluency removal tasks \cite{rasooli2013joint,honnibaljoint,tran2017joint}. Recently a transition-based neural model architecture was proposed for disfluency detection \cite{wang2017transition}. The current state of the art in disfluency detection \cite{wang2018semi} uses a neural machine translation framework with a transformer architecture and additional simulated data. All of the models mentioned above rely heavily on pattern match features, hand-crafted or automatically extracted, that help to identify repetitions and disfluencies with parallel syntactic structure.

While prosodic features are useful for detecting interruption points \cite{nakatani94,shriberg97,shriberg99,liu06}, recent methods on disfluency detection predominantly rely on lexical information exclusively. An exception is \cite{semi-markov}, which showed some gains using a simple concatenation of pause and word duration features. Similar to disfluency detection, parsing has seen little use of prosody in recent studies. However, \citet{tran2017joint} recently demonstrated that that a neural model using pause, word and rhyme duration, f0 and energy helps in spoken language parsing, specifically in the regions that contain disfluencies.

Early fusion and late fusion are the two most popular types of modality fusion techniques. In recent years, more interesting modality fusion approaches were introduced, most of them where the fusion happens inside the model \cite{zadeh2017tensor,chen2017multimodal,zadeh2018multi}. Those methods usually require the model to learn interactions between modalities implicitly, by backpropagating the errors based on the main objective function with respect to the task. Other multimodal representation learning approaches learn a shared representation between multiple modalities \cite{andrew2013deep,kiros2014unifying,xu2015show,suzuki2016joint}, often targeting unsupervised translation from one modality to the other. In our work we use innovations as a novel representation learning approach, where our emphasis is on looking into complementary cues rather than similarity between multiple modalities. 

%% file: conclusions.tex
\section{Conclusions}

In this paper, we introduce a novel approach to extracting acoustic-prosodic cues with the goal of improving disfluency detection, but also with the intention of impacting spoken language processing more generally. Our initial analysis of a text-only disfluency detection system shows that despite high performance of such models, there exists a big gap in the performance of text-based approaches for some types of disfluencies, such as restarts and non-trivial or long rephrases. Thus, prosody cues, which can be indicative of interruption points, have a potential to contribute towards detection of more difficult types of disfluencies. Since the acoustic-prosodic cues carry information related to multiple phenomena, it can be difficult to isolate the cues that are relevant to specific events, such as interruption points. In this work, we introduce a novel approach where we extract relevant acoustic-prosodic information using text-based distributional prediction of acoustic cues to derive vector z-score features, or innovations. The innovations point to irregularities in prosody flow that are not predicted by the text, helping to better isolate signals relevant to disfluency detection that are not simply redundant with textual cues. We explore both early and late fusion approaches to combine innovations with text-based features. Our experiments show that innovation features are better predictors of disfluencies compared to the original acoustic cues.   

Our analysis of the errors and of the innovation features point to a limitation of the current work, which is in the modeling of F0 features. The current model obtains word-based F0 (and energy) features by simply averaging the values over the duration of the word, which loses any distinctions between rising and falling F0. By leveraging polynomial contour models, we expect to improve both intonation and energy features, which we hope will reduce some of the false detections associated with emphasis and unexpected fluent phrase boundaries. 

An important next step is to test the system using ASR rather than hand transcripts.  It is possible that errors in the transcripts could hurt the residual prediction, but if prosody is used to refine the recognition hypothesis, this could actually lead to improved recognition. Finally, we expect that the innovation model of prosody can benefit other NLP tasks, such as sarcasm and intent detection, as well as detecting paralinguist information.




%% file: acknowledgements.tex
\subsection*{Acknowledgements}
We would like to thank anonymous reviewers
for their insightful comments. This work was funded in part by the US National Science Foundation, grant IIS-1617176.